\documentclass[10pt,twocolumn,letterpaper]{article}

\usepackage{cvpr}              % To produce the CAMERA-READY version
\usepackage[accsupp]{axessibility}  % Improves PDF readability for those with disabilities.
\usepackage{graphicx}
\usepackage{amsmath}
\usepackage{amssymb}
\usepackage{mathtools}
\usepackage{booktabs}
\usepackage{ulem}
\usepackage{colortbl}
\usepackage{arydshln}
\usepackage{vcell}
\usepackage{tabularray}
\usepackage{multirow}
\usepackage{arydshln}
\usepackage{adjustbox}
\usepackage{xcolor}
\usepackage{url}
\usepackage[pagebackref,breaklinks,colorlinks]{hyperref}

\usepackage[capitalize]{cleveref}
\crefname{section}{Sec.}{Secs.}
\Crefname{section}{Section}{Sections}
\Crefname{table}{Table}{Tables}
\crefname{table}{Tab.}{Tabs.}

\begin{document}

\title{How Close Are Other Computer Vision Tasks to Deepfake Detection?}

\author{Huy H. Nguyen$^1$, Junichi Yamagishi$^{1,2}$, and Isao Echizen$^{1,2,3}$ \\
$^1$National Institute of Informatics, Japan\ \ \ $^2$SOKENDAI, Japan\ \ \ $^3$The University of Tokyo, Japan\\
{\tt\small \{nhhuy, jyamagis, iechizen\}@nii.ac.jp}
}
\maketitle

%%%%%%%%% ABSTRACT
\begin{abstract}
   In this paper, we challenge the conventional belief that supervised ImageNet-trained backbones have strong generalizability and are suitable for use as feature extractors in deepfake detection models. We present a new measurement, \textit{``backbone separability,''} for visually and quantitatively assessing a backbone's raw capacity to separate data in an unsupervised manner. We also present a systematic benchmark for determining the correlation between deepfake detection and other computer vision tasks using backbones from pre-trained models. Our analysis shows that before fine-tuning, face recognition backbones are more closely related to deepfake detection than other backbones. Additionally, backbones trained using self-supervised methods are more effective in separating deepfakes than those trained using supervised methods. After fine-tuning all backbones on a small deepfake dataset, we found that self-supervised backbones deliver the best results, but there is a risk of overfitting. Our results provide valuable insights that should help researchers and practitioners develop more effective deepfake detection models.
\end{abstract}

%%%%%%%%% BODY TEXT
\section{Introduction}
\label{sec:intro}
Deepfake detection (DFD) has become an increasingly important task in recent years with the proliferation of high-quality generative models~\cite{tolosana2020deepfakes, rathgeb2022handbook}. DFD is a challenging task due to the sophistication of the generated content. As a result, most detection methods are deep-learning-based~\cite{malik2022deepfake} ones utilizing high-capacity models. However, the lack of large-scale and diverse deepfake datasets makes it difficult to effectively train models from scratch. To overcome this challenge, the use of transfer learning and fine-tuning has become a commonly used approach~\cite{rossler2019faceforensics++, nguyen2019capsule, ferrer2020deepfake, zhao2021multi, sun2022dual} in which researchers and practitioners leverage pre-trained backbones from computer vision (CV)-related models trained on ImageNet~\cite{deng2009imagenet}. In this way, the learned features from the pre-training can be transferred to the DFD task, making it possible to build more effective models even with limited training data. 

This paper challenges this conventional practice of using pre-trained image classification backbones for deepfake detection. While ImageNet has played a vital role in the success of several deep learning models, such as VGG~\cite{simonyan2014very}, ResNet~\cite{he2016deep}, XceptionNet~\cite{chollet2017xception}, and EfficientNet~\cite{tan2019efficientnet}, the effectiveness of the pre-trained backbones for the DFD task remains uncertain. Given that (facial) DFD is focused on the human face, we hypothesize that pre-trained backbones from facial-related tasks, such as face recognition and age estimation, may also be suitable. Additionally, the potential of self-supervised learning in learning effective representations suitable for various tasks has recently become recognized~\cite{schmarje2021survey}. Hence, there is a need for a systematic benchmark for evaluating the suitability of these backbones for the DFD task.

\begin{figure*}[t]
 \centering
 \includegraphics[width=120mm]{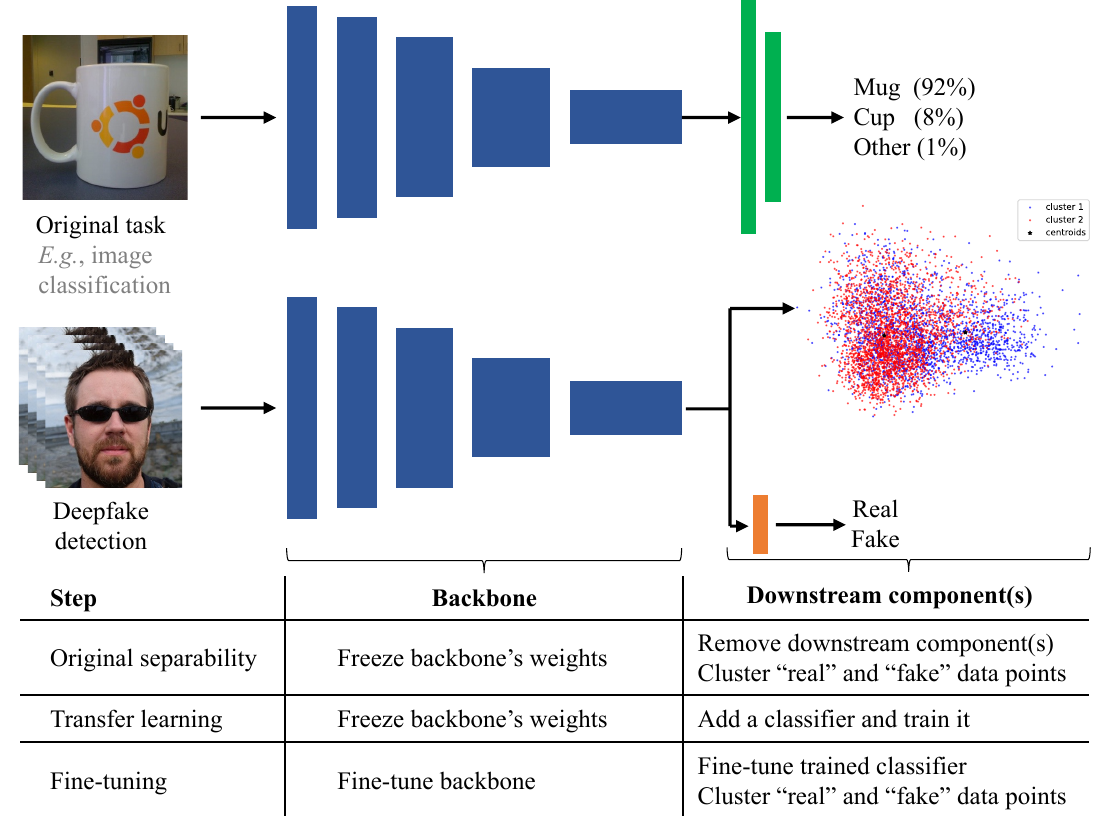}
 \caption{Overview of proposed systematic benchmark consisting of three steps: measuring the separability of pre-trained models, performing transfer learning, and fine-tuning. Best viewed in color.}
\label{fig:overview}
\end{figure*}

To ensure the accuracy of our systematic benchmark, we have established the following principles for conducting experiments in this paper:
\begin{itemize}
    \item \textit{Type of DFD models:} Since our benchmark focus on the backbones, we choose the most basic DFD approach, which takes images (video frames in the case of videos) as input. Such models from this approach consist of a backbone and a fully connected (FC) classifier. More complex models may utilize additional features or have complex modules, which may introduce nuisance factors making the benchmark inaccurate or biased. If a backbone demonstrates good separability on a simple model, it will likely perform well on more complex models.
    \item \textit{Type of deepfakes:} We focus on popular types of facial deepfakes~\cite{tolosana2020deepfakes}, comprising identity swaps, expression swaps, and entire face synthesis, whose datasets can be found or easily generated. We crop the faces from images or video frames during pre-processing.
    \item \textit{Datasets:} Since our objective is to measure the closeness between tasks, we use a small dataset for fine-tuning to assess whether the backbones can easily adapt to the DFD task and avoid catastrophic forgetting. The closeness measurement becomes meaningless if a model almost forgets its previously learned representations. A small dataset can also help reveal the model under-fitting or over-fitting, indicating that the original and DFD tasks are not closely related.
    \item \textit{Metrics:} Most benchmark methods for DFD rely on quantitative scores such as accuracy, equal error rate (EER), area under the curve (AUC), or half total error rate (HTER)~\cite{abdullakutty2021review}, which often lack explainability and rely heavily on the construction of the test set, especially when evaluating generalizability. Herefore, adding a more intuitive measurement is needed for DFD evaluation.
\end{itemize}

In summary, our contributions are threefold:
\begin{itemize}
 \item We present a new measurement, \textit{``backbone separability,''} for visually and quantitatively assessing a backbone's ability to separate data into pre-defined classes in an unsupervised manner using dimensional reduction and clustering.
 \item We present a systematic benchmark for measuring the similarity between DFD tasks and other CV tasks by evaluating the separability of various pre-trained and fine-tuned backbones in detecting different kinds of deepfakes.
 \item We demonstrate that fine-tuning and transfer learning with self-supervised backbones provide the best performance when dealing with a small deepfake dataset, although with a potential risk of overfitting. These insights can inform the development of more effective countermeasures against deepfakes, particularly for detecting unseen ones.
\end{itemize}

\section{Methodology}
\label{sec:med}

\subsection{Insights}
Finding the relationship between other machine learning tasks and deepfake detection is challenging and requires observing and measuring the changes in the pre-trained backbone before and after fine-tuning on DFD data. The complexity of deep learning and the curse of dimensionality make it necessary to filter out nuisance factors and simplify the problem. This can be achieved by using simple methods for the downstream module and employing dimensionality reduction algorithms to extract key features from the embeddings extracted by the backbone. We can infer the relationship between the tasks by evaluating the usefulness of these key features on DFD and observing their changes after fine-tuning.

\subsection{Benchmark Pipeline}
We propose using a systematic benchmark consisting of three steps to answer the question of how close other CV tasks are to DFD. The first step involves measuring the model's separability (details are discussed in section~\ref{sec:separability}), followed by evaluating transfer learning and fine-tuning (results are discussed in section~\ref{sec:experiment}). This process is illustrated in Fig.~\ref{fig:overview}. To ensure the accuracy of the measurement, it is essential to carefully construct a dataset (details are discussed in section~\ref{sec:dataset}) and select pre-trained models that cover a wide range and have a certain relationship for comparison (details are discussed in section~\ref{sec:models}).

In more detail, the first step involves obtaining feature extractors, also known as backbones, from the selected models. For instance, we eliminate the FC layers from the ResNet-50 model~\cite{he2016deep} pre-trained on the ImageNet dataset~\cite{deng2009imagenet}. We then assess their separability using part A of the prepared dataset. In the second step, we keep the backbone fixed, add an FC layer after it, and train the model using part B of the prepared dataset. In the final step, we unfreeze the backbone and fine-tune the entire model on part B. We also reassess the backbone's separability to examine the changes in the clusters within its embedding space. Both models from the second and final steps are evaluated using part C of the prepared dataset, and part D is used to test their generalizability. In the end, we analyze the results from these three steps to determine the extent of closeness of other CV tasks to DFD.

\begin{figure}[t]
 \centering
 \includegraphics[width=55mm]{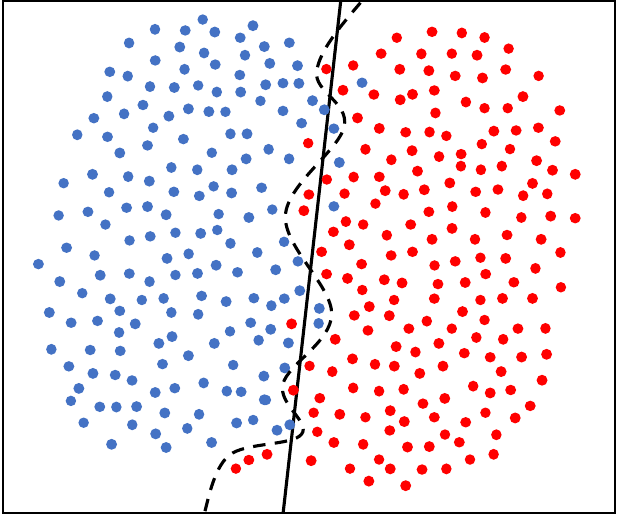}
 \caption{Decision boundaries of a high-capacity model (\textit{e.g.}, a neural network-based classifier), represented by a dashed curve, and a low-capacity model (\textit{e.g.}, a linear regression classifier), represented by a straight line. Best viewed in color.}
 \label{fig:separability}
\end{figure}

\subsection{Backbone Separability}
\label{sec:separability}
Machine learning models can generally be classified into two types: \textbf{generative} and \textbf{discriminative}~\cite{ng2001discriminative}. Generative models aim to model the data distribution, \textit{i.e.}, how the data is distributed throughout the space. On the other hand, discriminative models aim to learn the boundary that separates different classes of data in the feature space. Generative models are better suited for handling unseen data, while discriminative models are better suited for seen data. Most DFD models are discriminative and are the focus of this paper. With discriminative models, a relationship exists between model capacity (or capability) and overfitting. A high-capacity model has the ability to fit complex patterns but also noise and irrelevant patterns in the training data. An example of decision boundaries with a high-capacity model and a low-capacity one is visualized in Fig.~\ref{fig:separability}.

By utilizing the concepts of both types, we devised a measurement, $\mathcal{M}_{\mathcal{F_\theta}, \mathcal{R}_\phi, \mathcal{C}_\psi}: \mathbb{D} \mapsto \{\mathbb{E'}_i\}_{i=1}^{n}$, called \textit{\textbf{``backbone separability''}} for use in assessing a model's ability to separate data into pre-defined classes in an unsupervised manner. We first use the model's backbone, denoted as $\mathcal{F}_\theta$, to extract high-dimensional embeddings of all data points in a labeled dataset $\mathbb{D}$ representing the real-world distribution. Next, we use a dimensional reduction algorithm $\mathcal{R}_\phi$, such as principal component analysis (PCA)~\cite{pearson1901liii}, to reduce the embeddings $\mathbb{E}$ into 2D embeddings $\mathbb{E'}$. Then, a clustering algorithm $\mathcal{C}_\psi$, such as K-means~\cite{lloyd1982least}, is used to cluster $\mathbb{E'}$ into $n$ clusters $\{\mathbb{E'}_1...\mathbb{E'}_n\}$ (in the case of DFD, $n=2$). Finally, the clusters are visualized, and the accuracy is calculated using the labels in $\mathbb{D}$. The entire process is summarized in Equation~\ref{eq:separability}.

\begin{equation}
 \mathbb{E} = \mathcal{F}_\theta(\mathbb{D});\ \ \ \ \mathbb{E'} = \mathcal{R}_\phi(\mathbb{E});\ \ \ \ \{\mathbb{E'}_i\}_{i=1}^{n} = \mathcal{C}_\psi(\mathbb{E'})
 \label{eq:separability}
\end{equation}

The key insight behind this measurement is that it quantifies the ``raw'' separability of the model's backbone on a given task using the least-capacity unsupervised classifier in a reduced 2D embedding space. Good ``raw'' separation of the resulting clusters (yielding high separability accuracy) suggests that the backbone has effectively captured salient features and may exhibit good generalizability. On the other hand, poor separation indicates that the backbone is either unable to distinguish between different classes or is overfitted.

Both \textbf{cluster visualization} and \textbf{separation accuracy} are essential factors in assessing a backbone's separability. It is important to note that they do not always align with each other. For instance, in Fig.~\ref{fig:separability}, the two classes are well-separated \textit{visually} but only achievable with a high-capacity classifier, resulting in lower \textit{separation accuracy} by K-means, a low-capacity clustering method. This disagreement indicates that the backbone may learn some nuisance representations that harm its generalization. Conversely, if both classes are poorly visually separated and the separation accuracy is low, the proposed measurement suggests that the backbone cannot differentiate between deepfakes and genuine inputs. If both classes are well visually separated and the separation accuracy is high, it suggests that the backbone performs well on DFD.

\section{Experimental Design}
\label{sec:exp_design}
\subsection{Pre-trained Models}
\label{sec:models}

We selected several widely used CV models pre-trained on various datasets using different training methods and objective functions, including supervised and unsupervised learning. These models performed one of four tasks: face recognition, age estimation, image classification, and self-supervised learning. The selected models, their backbone architectures, and training datasets are listed in Table~\ref{tab:performances}. In addition to their original task, the pre-trained backbones of image classification models are widely used for various downstream tasks, including DFD. Self-supervised learning is not a specific task but a method of learning useful representations for other tasks~\cite{oord2018representation}.

We did not select models that perform the face detection task as they do not follow the common structure of the other tasks, making comparison difficult. For instance, models such as PyramidBox~\cite{tang2018pyramidbox} and RetinaFace~\cite{deng2020retinaface} rely on feature dynamics and have additional contextual modules (which other models do not have) that play a key role in their performance.

\begin{table*}
\centering
\caption{Performance overview of all models and random guess, sorted by backbone architecture, in terms of original backbone separability accuracies and EERs of transfer learning (TL) and fine-tuning (FT) models. DeiT III is excluded from the plots as it is an outlier.}
\label{tab:performances}
\adjustbox{max width=\textwidth}{
\begin{tabular}{l|l|l|l|c|c|c|c}
\multicolumn{1}{c|}{\textbf{Model name}} & \multicolumn{1}{c|}{\textbf{Backbone}} & \multicolumn{1}{c|}{\textbf{Task}} & \begin{tabular}[c]{@{}c@{}}\textbf{Backbone's}\\\textbf{pre-trained}\\\textbf{data}\end{tabular} & \begin{tabular}[c]{@{}c@{}}\textbf{Ori. backbone}\\\textbf{separability}\\\textbf{Acc. - set A$\uparrow$}\end{tabular} & \begin{tabular}[c]{@{}c@{}}\textbf{Ori. backbone}\\\textbf{separability}\\\textbf{Acc. - set C$\uparrow$}\end{tabular} & \begin{tabular}[c]{@{}c@{}}\textbf{TL}\\\textbf{model}\\\textbf{EER$\downarrow$}\end{tabular} & \begin{tabular}[c]{@{}c@{}}\textbf{FT}\\\textbf{model}\\\textbf{EER}$\downarrow$\end{tabular} \\ 
\hline
VGG-16~\cite{simonyan2014very} & VGG-16 & Image classification & ImageNet-1K & 52.38 & 52.59 & \textbf{22.17} & 14.37 \\
MWR (global)~\cite{shin2022moving} & VGG-16 & Age Estimation & UTK & \textbf{54.77} & 52.09 & 22.60 & \textbf{13.99} \\
\hdashline
ResNet-50~\cite{he2016deep} & ResNet-50 & Image classification & ImageNet-1K & 54.23 & 58.59 & 26.03 & 15.46 \\
BarlowTwins~\cite{zbontar2021barlow} & ResNet-50 & Self-supervised learning & ImageNet-1K & 53.37 & 59.33 & \textbf{21.16} & 14.09 \\
BYOL~\cite{grill2020bootstrap} & ResNet-50 & Self-supervised learning & ImageNet-1K & 54.85 & 60.50 & 21.25 & \textbf{13.61} \\
SimCLRv2~\cite{chen2020big} & ResNet-50 & Self-supervised learning & ImageNet-1K & \textbf{60.52} & \textbf{61.36} & 24.35 & 13.83 \\ 
\hdashline
iResNet-101~\cite{duta2021improved} & iResNet-101 & Image classification & ImageNet-1K & 52.93 & 58.83 & \textbf{21.86} & 14.17 \\
CosFace~\cite{wang2018cosface} & iResNet-101 & Face recognition & Glint360K & 54.60 & 62.23 & 30.39 & 14.45 \\
ArcFace~\cite{deng2019arcface} & iResNet-101 & Face recognition & MS-Celeb-1M & \textbf{64.80} & \textbf{65.03} & 31.08 & 14.70 \\
Partial FC~\cite{an2021partial} & iResNet-101 & Face recognition & Glint360K & 61.14 & 64.53 & 35.37 & \textbf{14.15} \\ 
\hdashline
FaceNet~\cite{schroff2015facenet} & Incep.-ResNet-v1 & Face recognition & VGGFace2 & 52.74 & 56.55 & 35.13 & 14.56 \\
Incep.-ResNet-v2~\cite{szegedy2017inception} & Incep.-ResNet-v2 & Image classification & ImageNet-1K & 52.05 & 57.22 & 31.83 & 14.35 \\
ResNet-101~\cite{he2016deep} & ResNet-101 & Image classification & ImageNet-1K & 52.56 & 58.49 & 26.98 & 16.72 \\
Xception~\cite{chollet2017xception} & XceptionNet & Image classification & ImageNet-1K & \textbf{53.80} & \textbf{59.33} & 29.04 & 14.24 \\
EfficientNet~\cite{tan2019efficientnet} & EfficientNet-B4 & Image classification & ImageNet-1K & 52.96 & 58.99 & 26.50 & 14.06 \\
EfficientNet-v2~\cite{tan2021efficientnetv2} & EfficientNetV2-M & Image classification & ImageNet-21K & 51.26 & 50.76 & \textbf{21.79} & \textbf{13.25} \\
\hdashline
\sout{DeiT III}~\cite{touvron2022deit} & DeiT III & Image classification & ImageNet-21K & 52.59 & 52.38 & 51.57 & 42.60 \\
\hdashline
Random guess & None & Random guess & None & 50.17 & 50.10 & 49.87 & 49.87   
\end{tabular}}
\end{table*}

\subsection{Datasets}
\label{sec:dataset}
This section introduces the datasets used to train the models for other CV tasks and the dataset used for our experiments.

\subsubsection{Datasets for training models for other CV tasks}

The CV models were pre-trained on various datasets by various authors. The model details and their corresponding training datasets are summarized in Table~\ref{tab:performances}. The details of each dataset are presented in Table~\ref{tab:pretrained_data}. Glint360K~\cite{an2021partial} is the largest dataset, while UTK~\cite{zhang2017age} is the smallest one. ImageNet-1K~\cite{deng2009imagenet} and ImageNet-21K~\cite{deng2009imagenet} contain the same images, but ImageNet-21K has 21 times the number of labels.

\subsubsection{Dataset used for experiments}

\begin{table*}[t]
\centering
\caption{Detailed information of datasets used for training selected CV models.}
\label{tab:pretrained_data}
% \adjustbox{max width=\columnwidth}{
\begin{tabular}{l|c|c|r|rl}
\multicolumn{1}{c|}{\textbf{Dataset name}} & \textbf{Main task} & \textbf{Year} & \multicolumn{1}{c|}{\begin{tabular}[c]{@{}c@{}}\textbf{Size}\\\textbf{(approx.)}\end{tabular}} & \multicolumn{2}{c}{\textbf{Remark}} \\ 
\hline
ImageNet-1K~ & Image classification & 2012 & 1,430,000 & 1,000 & classes \\
ImageNet-21K~ & Image classification & 2021 & 1,430,000 & 21,000 & classes \\
UTK~ & Age estimation & 2017 & 20,000 & 0 - 116 & years old \\
MS-Celeb-1M~ & Face recognition & 2016 & 10,000,000 & 100,000 & identities \\
VGGFace2~ & Face recognition & 2018 & 3,300,000 & 9,000 & identities \\
Glint360K~ & Face recognition & 2021 & 17,000,000 & 360,000 & identities
\end{tabular}
% }
\end{table*}

\begin{table}[t]
\centering
\caption{Number of real and fake images of three subsets.}
\label{tab:proposed_data}
\begin{tabular}{c|c|c|c}
\textbf{Name} & \textbf{Purpose} & \textbf{\begin{tabular}[c]{@{}c@{}}Number of\\ real images\end{tabular}} & \textbf{\begin{tabular}[c]{@{}c@{}}Number of\\ fake images\end{tabular}} \\ \hline
Set A & Clustering & 44,037 & 55,963\\
Set B & Training & 13,200 & 13,000\\
Set C & Testing (seen) & 10,000 & 11,000\\
Set D & Testing (unseen) & 200 & 200
\end{tabular}
\end{table}

We gathered facial images and videos from various sources to construct the dataset for our experiments. Our dataset consists of four subsets (A, B, C, and D), as detailed in Table~\ref{tab:proposed_data}. We balanced the subsets regarding the ratio of real and fake images and the number of images per training method. Furthermore, subsets A, B, C, and D are locally mutually exclusive, guaranteeing no overlap between them from each source dataset. Subset A, the largest one, was used to measure the separability of the pre-trained models, subset B was used for training, and subsets C and D were used for evaluation.

Regarding the normal subsets (A, B, and C), for the \textbf{real} part, we gathered images from the VidTIMIT dataset~\cite{sanderson2009multi}, VoxCeleb2 dataset~\cite{chung2018voxceleb2}, FaceForensics++ (FF++) dataset~\cite{rossler2019faceforensics++}, Google DFD dataset~\cite{googledfd}, Deepfake Detection Challenge Dataset (DFDC)~\cite{dolhansky2020deepfake}, and the Celeb-DF dataset~\cite{li2020celeb}. For the \textbf{fake} part, we gathered images from the FF++ dataset, Google DFD dataset, Celeb-DF dataset, DFDC, DeepfakeTIMIT (DF-TIMIT) dataset~\cite{korshunov2018deepfakes}, and YouTube-DF dataset~\cite{kukanov2020cost}. We also generated several generative adversarial network (GAN) images using StarGAN~\cite{choi2018stargan}, StarGAN-v2~\cite{choi2020stargan}, RelGAN~\cite{wu2019relgan}, ProGAN~\cite{karras2018progressive}, StyleGAN~\cite{karras2019style}, and StyleGAN2~\cite{karras2020analyzing}.

Regarding the special subsets used for generalizability evaluation (subset D), for the \textbf{real} part, we gathered images from the Glint360K~\cite{an2021partial} dataset. For the \textbf{fake} part, we collected images from the dataset constructed by Afchar \textit{et al.}~\cite{afchar2018mesonet} (which mainly contains cropped faces from pornography videos) and facial images generated using a latent diffusion model trained on facial images~\cite{rombach2022high}.

\subsection{Metrics}

In DFD, accuracy, EER, AUC, and HTER are commonly used metrics. However, EER and AUC cannot be used in backbone separability measurement, which relies on clustering. As the clusters are formed unsupervised, it is impossible to distinguish between ``real'' and ``fake'' ones. Therefore, if the accuracy is less than 50\%, we have to reverse the assumed classes. To further investigate a backbone's inclination towards real or deepfakes, we use true positive rate (TPR) and true negative rate (TNR). While other extrinsic and intrinsic measures for clustering are available, we limit ourselves to these metrics for simplicity.

When evaluating transfer-learned and fine-tuned models (consisting of a backbone and a classifier), there is no validation set available for classification threshold calibration, as only training set B and test sets C and D are provided. Therefore, EER is the most suitable metric to measure a model's performance. We also use HTER with a threshold of 0.5 on set D to observe the changes in distributions in the embedding spaces after fine-tuning.

\subsection{Evaluation Settings}
For measuring backbone separability, we used PCA~\cite{pearson1901liii} for dimensionality reduction and K-means~\cite{lloyd1982least} for clustering. We trained the model for 50 epochs for transfer learning and selected the checkpoint with the lowest EER on set B. Next, we fine-tuned the selected transfer-learned model for 400 epochs. As large models may require extended training time to converge and small models may converge quickly but easily lead to overfitting afterward, we tested checkpoints 100 and 400 of each model on set C. We reported the better result among the two.

\section{Experimental Results and Discussions}
\label{sec:experiment}

\subsection{Overview}
\begin{figure}[t]
 \centering
 \includegraphics[width=\columnwidth]{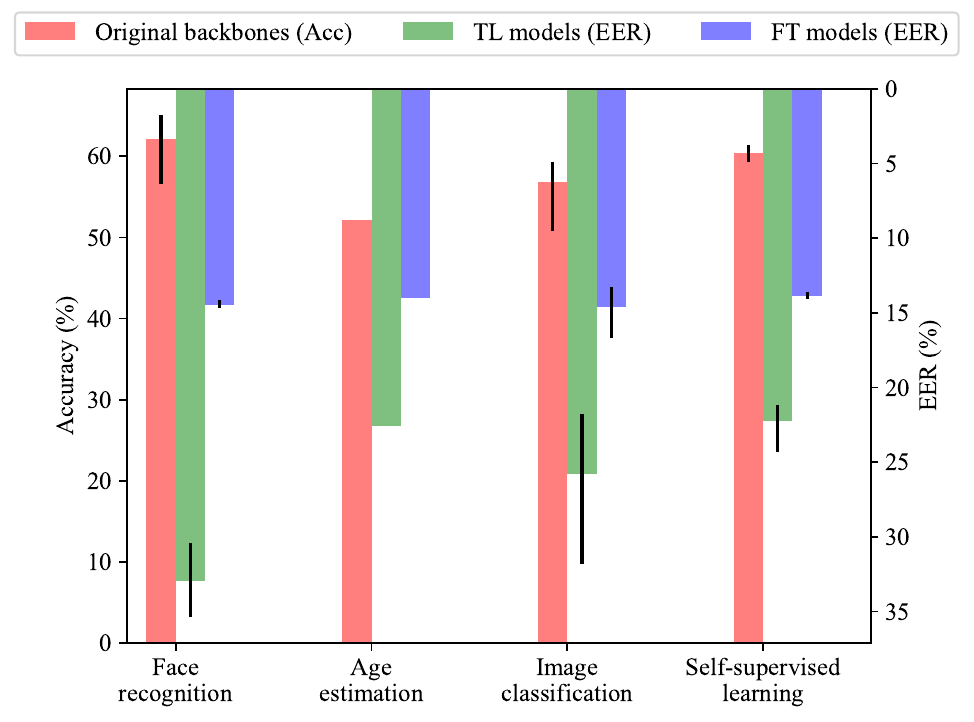}
 \caption{Comparison of model performances based on tasks. Best viewed in color.}
\label{fig:comp_task}
\end{figure}

\begin{figure}[t]
 \centering
 \includegraphics[width=\columnwidth]{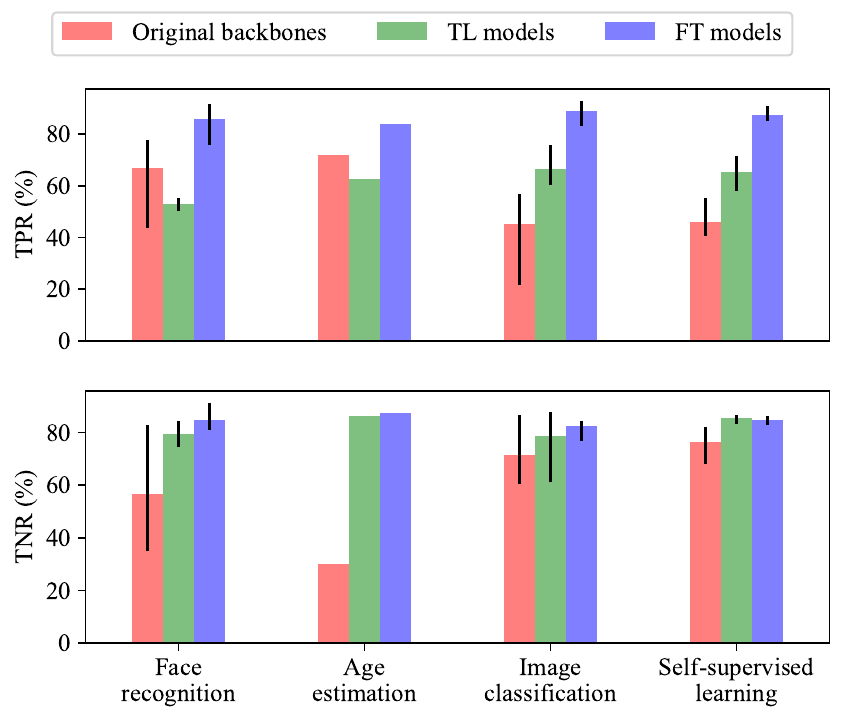}
 \caption{Detailed performances of models based on tasks in terms of TPR and TNR. Best viewed in color.}
\label{fig:comp_tprtnr}
\end{figure}

The results of the experiments are summarized in Table~\ref{tab:performances}, sorted by backbone architecture. We acknowledge that some architectures are only present in one task due to the absence of pre-trained models, which makes it difficult to compare them fairly. Therefore, we only compared architectures that appeared in more than one task. The results for single-present architectures were used as reference points. Notably, the DeiT III~\cite{touvron2022deit} backbone, which is a high complexity vision transformer, could not work properly after transfer learning and fine-tuning on the small DFD dataset. Thus, we excluded it from the four figures  (Figures~\ref{fig:comp_task},~\ref{fig:comp_tprtnr},~\ref{fig:comp_arc},  and~\ref{fig:comp_detail}) in this section.

Regarding the question of how close other CV tasks are to DFD, the results suggest that backbones from face recognition tasks, such as ArcFace and Partial FC, as well as self-supervised learning techniques like SimCLRv2, show better separability in DFD than the other backbones. These findings support our hypothesis that face recognition tasks are closely related to DFD and that certain self-supervised learning techniques can learn representations useful for distinguishing real from fake. In contrast, pre-trained backbones trained on ImageNet data were less effective in separating real from fake.

\subsection{Results in Detail}

\subsubsection{Changes in backbone separability}

\begin{figure*}[t]
 \centering
 \includegraphics[width=\textwidth]{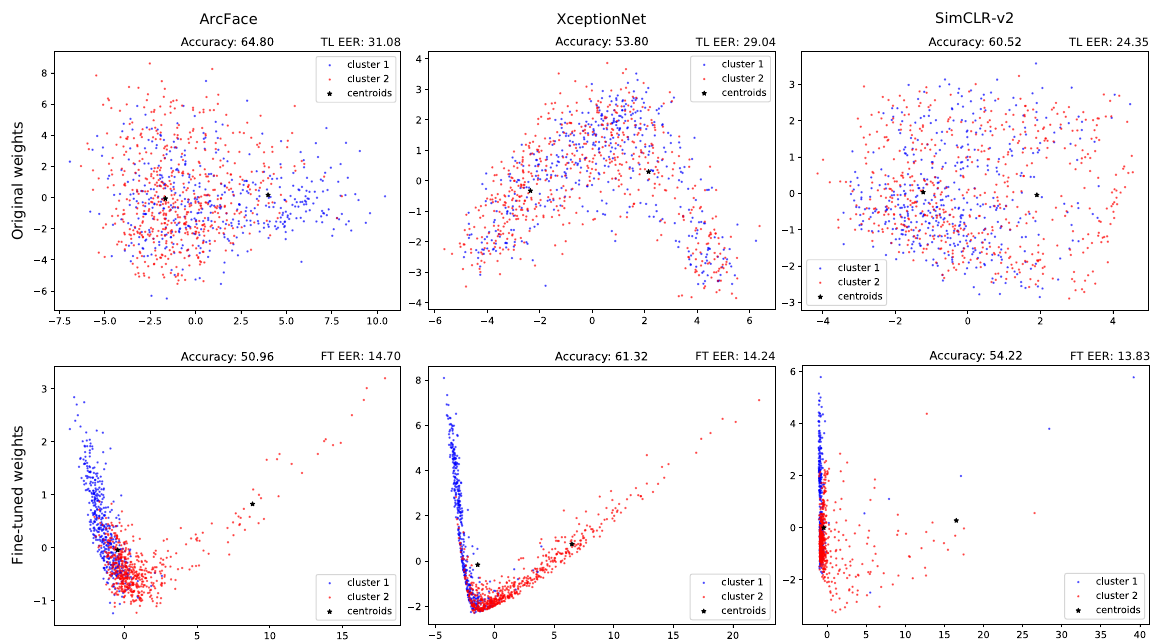}
 \caption{Visualization of clusters in the embedding spaces and separability accuracies obtained using proposed measurement on original and fine-tuned backbones of three widely used models: ArcFace, XceptionNet, and SimCLR-v2. One thousand data points were randomly selected for visualization. The EERs are presented for both transfer learning and fine-tuning scenarios. The former involves using the backbone without any changes to its weights while the latter involves modifying the weights of the backbone. Best viewed in color.}
\label{fig:embeddings}
\end{figure*}

Figure~\ref{fig:embeddings} displays the outcomes of our proposed measurement on the original and fine-tuned backbones of three widely used models: ArcFace, XceptionNet, and SimCLR-v2. Prior to fine-tuning, both ArcFace's and SimCLR-v2's embedding spaces exhibit some areas that contain nearly real or fake data points, suggesting that these pre-trained backbones have some degree of deepfake separability. However, this phenomenon is hard to observe in XceptionNet's embedding space. The separation accuracy also shows a similar trend with 64.80\% and 60.52\% for ArcFace and SimCLR-v2, respectively, and only 53.80\% for XceptionNet.

After fine-tuning, all models showed improved separability in terms of visualization. However, the cluster shapes made it challenging for a low-capacity classifier to differentiate between them. Consequently, the unsupervised clustering algorithm K-means failed to cluster them accurately. The reduced 2D space still showed some overlap between the two classes, which indicates that the models might learn not only DFD representations but also nuisance ones. These results imply that while fine-tuning can improve the models' DFD detection capability, there is a potential risk of overfitting, which could negatively impact generalizability. (discussed in Section~\ref{sec:unseen}).

\subsubsection{Comparison by task}
Table~\ref{tab:performances} and Fig.~\ref{fig:comp_task} demonstrate that facial-task-trained backbones generally offer better pre-trained separability than backbones trained on other tasks, such as image classification. Interestingly, moving window regression (MWR), an age estimation backbone with poor pre-trained separability, achieved good performance after transfer learning and fine-tuning. Self-supervised learning backbones, however, performed best overall in all stages. Even with state-of-the-art architectures like XceptionNet, EfficientNet, and EfficientNet-v2 included in the calculation of the performance statistics, the image classification backbones achieved only average overall performance.

In addition to accuracy and EER, we also measured the TPR and TNR, which are shown in Figure~\ref{fig:comp_tprtnr}. We found that the pre-trained face recognition and age estimation backbones were more prone to identifying fake images, resulting in higher TPR but lower TNR. On the other hand, after fine-tuning, the image classification and self-supervised learning backbones were more sensitive to fake images, leading to higher TPR.

\begin{figure*}[t]
 \centering
 \includegraphics[width=130mm]{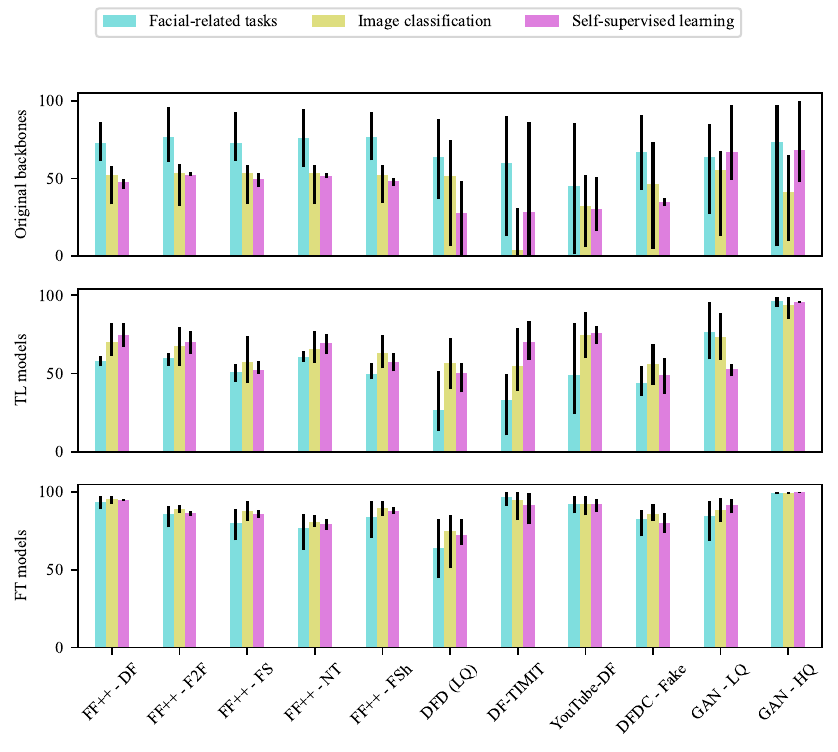}
 \caption{Detailed accuracies of models on various deepfake methods in three stages of the proposed benchmark. Facial-related tasks comprise age estimation and face recognition tasks. LQ and HQ stand for low quality and high quality, respectively. Best viewed in color.}
\label{fig:comp_detail}
\end{figure*}

We also examined the accuracies of the models for each deepfake method and averaged them by the models' original tasks. As shown in Fig.~\ref{fig:comp_detail}, the highly compressed Google DFD dataset was the most challenging, followed by the DFDC datasets. Interestingly, images created by Thies \textit{et al.} 's neural texture method~\cite{thies2019deferred} became more difficult to detect after fine-tuning, suggesting that it has characteristics different from those of the other deepfake methods.

\begin{figure}[t]
 \centering
 \includegraphics[width=\columnwidth]{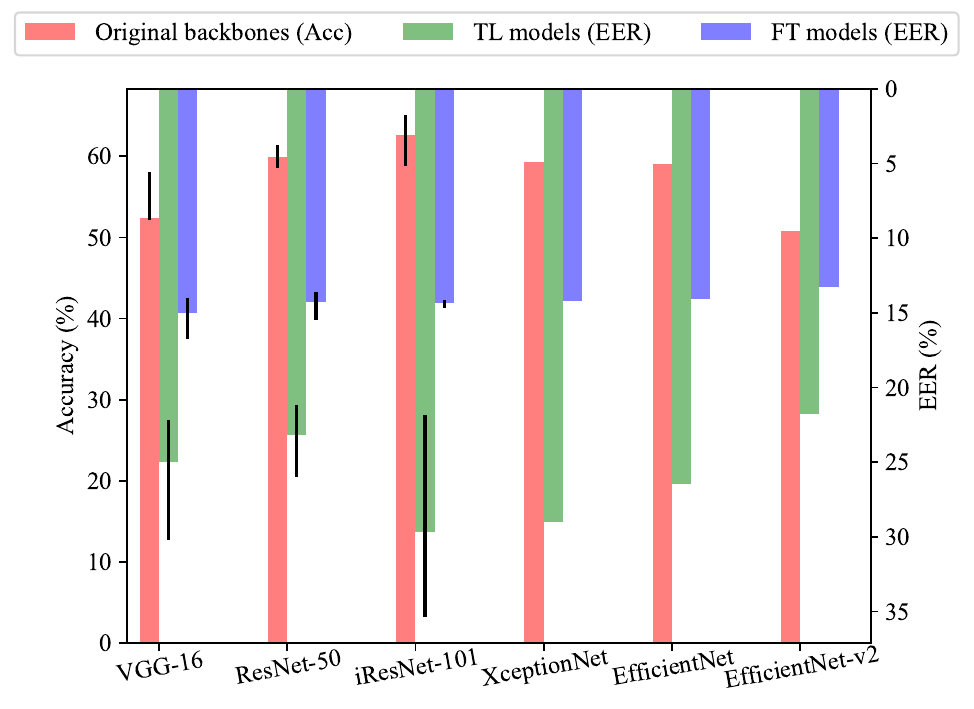}
 \caption{Comparison of model performances based on backbone architecture. Best viewed in color.}
\label{fig:comp_arc}
\end{figure}

\subsubsection{Comparison by architecture}

While comparison by architecture may not directly address the question posed in the title of our paper, it can still provide valuable insights into the performance of various deep-learning architectures for DFD. As shown in Fig.~\ref{fig:comp_arc}, modern architectures such as XceptionNet, EfficientNet, and EfficientNet-v2 consistently outperformed the other architectures after fine-tuning. This may help to explain why they are frequently used in DFD models\cite{rossler2019faceforensics++, ferrer2020deepfake, zhao2021multi, sun2022dual}. Notably, the original pre-trained EfficientNet-v2 achieved the lowest backbone separability accuracy. However, it outperformed all other backbones by a certain margin after fine-tuning. This finding suggests that self-supervised training with EfficientNet-v2 (no pre-trained model available) may lead to better representations than with ResNet-50.

\subsubsection{Comparison by training data}
After linking Table~\ref{tab:pretrained_data} and Table~\ref{tab:performances}, we observed that the size of the dataset and the number and detail of annotations used in pre-training are essential factors in pre-trained backbones' performance. For instance, pre-training with Glint360K, the largest dataset, helped the Partial FC backbone to perform better than the CosFace and ArcFace backbones after being fine-tuned on the DFD dataset. It is also possible that the 21K labels in the ImageNet dataset contributed to the high performance of the EfficientNet-v2 backbone, but this remains a hypothesis and requires further investigation.

\subsubsection{Generalizability}
\label{sec:unseen}

We evaluated the performance of the original and fine-tuned backbones of the three models mentioned in Fig.~\ref{fig:embeddings} on set D using accuracy, EER, AUC, and HTER (with a standard classification threshold of 0.5 for HTER). The results are presented in Table~\ref{tab:unseen} and indicate that all models exhibited lower performances on unseen data than on seen data. Although the detectors successfully identified images generated by latent diffusion models, they struggled to detect deepfake pornography images, and the fine-tuned models often misclassified real images as fake. As hypothesized, the fine-tuned models, with their complex distributions in the embedding spaces, performed worse on unseen data than transfer-learned models that retained their original backbone weights. The significant shifts in EER thresholds and noticeable increases in HTERs suggest that the fine-tuned classifiers' decision boundaries differed substantially between seen and unseen data.

\begin{table*}[t]
\centering
\caption{Accuracies, EERs with their corresponding thresholds, AUC, and HTERs of transfer-learned (TL) backbones and fine-tuned (FT) backbones on unseen test set D. A threshold of 0.5 was used to calculate accuracies and HTERs.}
\label{tab:unseen}
% \adjustbox{max width=\columnwidth}{
\begin{tabular}{l|c|lcc|c|c|c|c}
\multicolumn{1}{c|}{\multirow{3}{*}{\textbf{Model name }}} & \multirow{3}{*}{\textbf{Step}} & \multicolumn{3}{c|}{\textbf{Accuracy}} & \multirow{3}{*}{\textbf{EER}} & \multirow{3}{*}{\begin{tabular}[c]{@{}c@{}}\textbf{EER}\\\textbf{threshold}\end{tabular}} & \multirow{3}{*}{\textbf{AUC}} & \multirow{3}{*}{\textbf{HTER}} \\ 
\cline{3-5}
\multicolumn{1}{c|}{} &  & \multicolumn{1}{c}{\textbf{Real}} & \begin{tabular}[c]{@{}c@{}}\textbf{Latent}\\\textbf{diffusion}\end{tabular} & \begin{tabular}[c]{@{}c@{}}\textbf{Porn}\\\textbf{DF}\end{tabular} &  &  &  \\ 
\hline
\multirow{3}{*}{ArcFace} & TL & 68.50 & 100.00 & 45.00 & 28.50 & 0.5043 & 81.84 & 29.50 \\
 & FT & 19.50 & 99.00 & 76.00 & 31.00 & 0.9989 & 72.71 & 46.50 \\ 
\hdashline
\multirow{2}{*}{XceptionNet} & TL & 55.00 & 100.00 & 48.00 & 33.00 & 0.5781 & 72.63 & 35.50 \\
 & FT & 44.50 & 100.00 & 48.00 & 34.50 & 0.9492 & 72.10 & 40.75 \\ 
\hdashline
\multirow{2}{*}{SimCLR-v2} & TL & 69.00 & 99.00 & 35.00 & 32.50 & 0.4932 & 73.60 & 32.00 \\
 & FT & 26.50 & 98.00 & 46.00 & 39.00 & 0.9815 & 64.71 & 50.75
\end{tabular}
% }
\end{table*}

\section{Conclusion}
With our proposed systematic benchmark and intuitive measurement, we have demonstrated that using the backbones of supervised ImageNet-trained models for DFD is not the optimal choice. Our results show that with the same architecture, backbones trained for facial tasks and ones trained with self-supervised learning techniques offer better performance, with the latter being the best overall. Additionally, we found that the size of the dataset and the number of annotations are also important factors to consider when evaluating pre-trained backbones. Although we could not perform a comprehensive and truly fair comparison, our findings provide valuable insights for developing more effective countermeasures against deepfakes, especially for detecting unseen ones.

In future work, we suggest exploring the regularization of the backbones during fine-tuning to increase their raw separability and thereby improve their generalizability. Additionally, we suggest exploring the feasibility of training DFD models using self-supervised learning or hybrid approaches that incorporate multi-task learning.

\section*{Acknowledgements}
This work was partially supported by JSPS KAKENHI Grants JP18H04120, JP20K23355, JP21H04907, and JP21K18023, and by JST CREST Grants JPMJCR18A6 and JPMJCR20D3, including the AIP challenge program, Japan.

%%%%%%%%% REFERENCES
{\small
\bibliographystyle{ieee_fullname}
\bibliography{refs}
}

\end{document}